\def\year{2017}\relax
\documentclass[letterpaper]{article}
\usepackage{aaai17}
\usepackage{times}
\usepackage{helvet}
\usepackage{courier}
\usepackage[pdftex]{graphicx}
\usepackage{amsmath}
\usepackage{url}
\usepackage[linesnumbered,ruled]{algorithm2e}
\usepackage{multirow}
\usepackage{txfonts}
\usepackage{centernot}
\usepackage{caption}
\usepackage{subcaption}
\usepackage{comment}
\usepackage{booktabs}
\usepackage{makecell}

\newtheorem{definition}{Definition}
\newtheorem{theorem}{Theorem}

\begin{document}
%
\title{A Causal Framework for Discovering and Removing \\ Direct and Indirect Discrimination}
\author{Lu Zhang, Yongkai Wu, and Xintao Wu\\
University of Arkansas\\
\{lz006,yw009,xintaowu\}@uark.edu}
\maketitle
\begin{abstract}
Anti-discrimination is an increasingly important task in data science. 
In this paper, we investigate the problem of discovering both direct and indirect discrimination from the historical data, and removing the discriminatory effects before the data is used for predictive analysis (e.g., building classifiers). We make use of the causal network to capture the causal structure of the data. Then we model direct and indirect discrimination as the \emph{path-specific effects}, which explicitly distinguish the two types of discrimination as the causal effects transmitted along different paths in the network. Based on that, we propose an effective algorithm for discovering direct and indirect discrimination, as well as an algorithm for precisely removing both types of discrimination while retaining good data utility. Different from previous works, our approaches can ensure that the predictive models built from the modified data will not incur discrimination in decision making. Experiments using real datasets show the effectiveness of our approaches.
\end{abstract}

\section{Introduction}

Discrimination refers to unjustified distinctions in decisions against individuals based on their membership in a certain group. Federal Laws and regulations (e.g., the Equal Credit Opportunity Act of 1974) have been established to prohibit discrimination on several grounds, such as gender, age, sexual orientation, race, religion or belief, and disability or illness, which are referred to as the \emph{protected attributes}. Nowadays various predictive models have been built around the collection and use of historical data to make important decisions like employment, credit and insurance. If the historical data contains discrimination, the predictive models are likely to learn the discriminatory relationship present in the data and apply it when making new decisions. Therefore, it is imperative to ensure that the data goes into the predictive models and the decisions made with its assistance are not subject to discrimination.

In the legal field, discrimination is usually divided into direct and indirect discrimination. Direct discrimination occurs when individuals receive less favorable treatment explicitly based on the protected attributes. An example would be rejecting a qualified female applicant in applying a university just because of her gender. Indirect discrimination refers to the situation where the treatment is based on apparently neutral non-protected attributes but still results in unjustified distinctions against individuals from the protected group. A well-known example of indirect discrimination is redlining, where the residential Zip code of the individual is used for making decisions such as granting a loan. Although Zip code is apparently a neutral attribute, it correlates with race due to the racial makeups of certain areas. Thus, Zip code can indirectly lead to racial discrimination if there is no good reason to justify its use in making decisions.

Discrimination discovery and removal has received an increasing attention over the past few years in data science \cite{hajian2013methodology,kamiran2012data,ruggieri2010data,romei2014multidisciplinary,feldman2015certifying}. Many approaches have been proposed to deal with both direct and indirect discrimination. However, significant issues exist in current techniques. For discrimination discovery, the difference in decisions across the protected and non-protected groups is a combined (not necessarily linear) effect of direct discrimination, indirect discrimination, and other effects which are objectively explainable and should not be considered as discrimination. However, few works have explicitly identified the three different effects when measuring discrimination. For example, the classic metrics \emph{risk difference}, \emph{risk ratio}, \emph{relative chance}, \emph{odds ratio}, etc. \cite{romei2014multidisciplinary} treat all the difference in decisions as discrimination. \cite{zliobaite2011handling} considered the explainable effect, but failed to distinguish the effects of direct and indirect discrimination. For discrimination removal, an algorithm must ensure that the predictive models built from the historical data do not incur discrimination in decision making. However, as we will show in our experiments, previous works in removing discrimination cannot guarantee that the predictive models are not subject to discrimination even though they attempt to modify the historical data to contain no discrimination. In addition, it is a general requirement is to preserve the data utility while achieving non-discrimination. As will also shown in the experiments, totally removing all connections between the protected attribute and decision as proposed in \cite{feldman2015certifying} may suffer significant utility loss.



\begin{figure}
	\centering
		\includegraphics[width=1.5in]{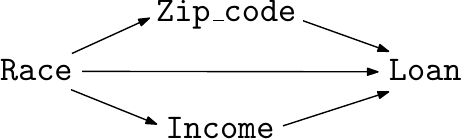}
	\caption{The toy model.}
	\label{fig:toy}
\end{figure}

The causal modeling based discrimination detection has been proposed most recently \cite{DBLP:journals/corr/BonchiHMR15,zhang2016situation,zhang2016discrimination} for improving the correlation based approaches. In this paper, we develop a framework for discovering and removing both direct and indirect discrimination based on the causal network. A causal network is a directed acyclic graph (DAG) widely used for causal representation, reasoning and inference \cite{pearl2009causality}, where causal effects are carried by the paths that trace arrows pointing from the cause to the effect which are referred to as the \emph{causal paths}. Using this model, direct and indirect discrimination can be captured by the causal effects of the protected attribute on the decision transmitted along different paths. Direct discrimination is modeled by the causal effect transmitted along the direct path from the protected attribute to the decision. Indirect discrimination, on the other hand, is modeled by the causal effect transmitted along other causal paths that contain any unjustified attribute. Consider a toy model of a loan application system shown in Figure \ref{fig:toy} for example. Assume that we treat \texttt{Race} as the protected attribute, \texttt{Loan} as the decision, and \texttt{Zip\_code} as the unjustified attribute that causes redlining. Direct discrimination is then modeled by path $\texttt{Race}\rightarrow \texttt{Loan}$, and indirect discrimination is modeled by path $\texttt{Race}\rightarrow \texttt{Zip\_code}\rightarrow \texttt{Loan}$. Assume that the use of $\texttt{Income}$ can be objectively justified as it is reasonable to deny a loan if the applicant has low income. In this case, path $\texttt{Race}\rightarrow \texttt{Income}\rightarrow \texttt{Loan}$ is explainable, which means that the difference in loan issuance across different race groups can be explained by the fact that some race groups in the dataset tend to be under-paid.

To measure the causal effect transmitted along a certain causal path, we employ the formulation of the \emph{path-specific effect} \cite{avin2005identifiability,shpitser2013counterfactual}. We define direct and indirect discrimination as different path-specific effects and show how to accurately measure them using the observational data. Based on that, we propose an effective algorithm for discovering direct and indirect discrimination, as well as an algorithm for precisely removing both types of discrimination while retaining good data utility. Our approaches can ensure that the predictive models built from the modified data are not subject to any type of discrimination. The experiments using two real datasets show that our approaches are effective in discovering and removing discrimination.


\section{Preliminary Concepts}

A causal network is a DAG $\mathcal{G}=(\mathbf{V},\mathbf{A})$ where $\mathbf{V}$ is a set of nodes and $\mathbf{A}$ is a set of arcs. Each node represents an attribute. Each arc, denoted by an arrow $\rightarrow$ pointing from the cause to the effect represents the direct causal relationship. Throughout the paper, we denote an attribute by an uppercase alphabet, e.g., $X$; denote a subset of attributes by a bold uppercase alphabet, e.g., $\mathbf{X}$. We denote a domain value of attribute $X$ by a lowercase alphabet, e.g., $x$; denote a value assignment of attributes $\mathbf{X}$ by a bold lowercase alphabet, e.g., $\mathbf{x}$. For a node $X$, its parents are denoted by $Pa(X)$, and its children are denoted by $Ch(X)$. Each node is associated with a conditional probability table (CPT), i.e., $P(x|Pa(X))$. We also use $Pa(X)$ to represent a value assignment of $X$'s parents if no unambiguity occurs in the context. The joint distribution over all attributes $P(\mathbf{v})$ can be computed using the factorization formula \cite{koller2009probabilistic}
\begin{equation}\label{eq:bn}
P(\mathbf{v}) = \prod_{V\in \mathbf{V}}P(v|Pa(V)),
\end{equation}
where $P(v|Pa(V))$ is the observational distribution.

In the causal network, the measuring of causal effects is facilitated with the $do$-calculus \cite{pearl2009causality}, which simulates the physical interventions that force some attributes $\mathbf{X}$ to take certain values $\mathbf{x}$. The post-intervention distributions, which represent the effect of the intervention, can be estimated from the observational data. Formally, the intervention that sets the value of $\mathbf{X}$ to $\mathbf{x}$ is denoted by $do(\mathbf{X} = \mathbf{x})$. The post-intervention distribution of all other attributes $\mathbf{Y}=\mathbf{V}\backslash \mathbf{X}$, i.e., $P(\mathbf{Y}=\mathbf{y}|do(\mathbf{X}=\mathbf{x}))$ or simply $P(\mathbf{y}|do(\mathbf{x}))$, can be expressed as a truncated factorization formula \cite{pearl2009causality}
\begin{equation*}
P(\mathbf{y}|do(\mathbf{x})) = \prod_{Y\in \mathbf{Y}}P(y|Pa(Y))\delta_{\mathbf{X}=\mathbf{x}},
\end{equation*}
where $\delta_{\mathbf{X}=\mathbf{x}}$ means assigning any attributes in $\mathbf{X}$ involved in the term ahead with the corresponding values in $\mathbf{x}$. Specifically, the post-intervention distribution of a single attribute $Y$ given an intervention on a single attribute $X$ is given by
\begin{equation}\label{eq:do}
P(y|do(x)) = \sum_{\mathbf{V}\backslash \{X,Y\},Y=y}\prod_{V\in \mathbf{V}\backslash \{X\}}P(v|Pa(V))\delta_{X=x},
\end{equation}
where the summation is a marginalization that traverses all value combinations of $\mathbf{V}\backslash \{X,Y\}$. 

By using the $do$-calculus, the total causal effect of $X$ on $Y$ is defined as follows \cite{pearl2009causality}. Note that in this definition, the effect of the intervention is transmitted along all causal paths from the cause $X$ to the effect $Y$.
\begin{definition}[Total causal effect]
The total causal effect of the change of $X$ from $x_{1}$ to $x_{2}$ on $Y=y$ is given by
\begin{equation*}
TE(x_{2},x_{1}) = P(y|do(x_{2}))-P(y|do(x_{1})).
\end{equation*}
\end{definition}

The path-specific effect is an extension to the total causal effect in the sense that the effect of the intervention is transmitted only along a subset of causal paths from $X$ to $Y$. Denote a subset of causal paths by $\pi$. The $\pi$-specific effect considers a counterfactual situation where the effect of $X$ on $Y$ with the intervention is transmitted along $\pi$, while the effect of $X$ on $Y$ without the intervention is transmitted along paths not in $\pi$. We denote by $P(y\mid do(x_{2}|_{\pi}))$ the distribution of $Y$ after an intervention of changing $X$ from $x_{1}$ to $x_{2}$ with the effect transmitted along $\pi$. Then, the $\pi$-specific effect of $X$ on $Y$ is defined as follows \cite{avin2005identifiability}.
\begin{definition}[Path-specific effect]
Given a path set $\pi$, the $\pi$-specific effect of the value change of $X$ from $x_{1}$ to $x_{2}$ on $Y=y$ is given by
\begin{equation*}
SE_{\pi}(x_{2},x_{1}) = P(y\mid do(x_{2}|_\pi)) - P(y\mid do(x_{1})).
\end{equation*}
\end{definition}


The authors in \cite{avin2005identifiability} have given the condition under which the path-specific effect can be estimated from the observational data, known as the recanting witness criterion.

\begin{definition}[Recanting witness criterion]\label{def:rwc}
Given a path set $\pi$, let $Z$ be a node in $\mathcal{G}$ such that: 1) there exists a path from $X$ to $Z$ which is a segment of a path in $\pi$; 2) there exists a path from $Z$ to $Y$ which is a segment of a path in $\pi$; 3) there exists another path from $Z$ to $Y$ which is not a segment of any path in $\pi$. Then, the recanting witness criterion for the $\pi$-specific effect is satisfied with $Z$ as a witness.
\end{definition}
\begin{theorem}[Identifiability]\label{thm:rwc}
The $\pi$-specific effect can be estimated from the observational data if and only if the recanting witness criterion for the $\pi$-specific effect is not satisfied.
\end{theorem}

If the recanting witness criterion is not satisfied, the $\pi$-specific effect $SE_{\pi}(x_{2},x_{1})$ can be computed as follows based on \cite{shpitser2013counterfactual}. First, express $P(y|do(x_{1}))$ as the truncated factorization formula according to Equation \eqref{eq:do}. Second, to compute $P(y\mid do(x_{2}|_{\pi}))$, divide the children of $X$ into two sets $\mathbf{S}_{\pi}$ and $\bar{\mathbf{S}}_{\pi}$, i.e., $Ch(X) = \mathbf{S}_{\pi}\cup \bar{\mathbf{S}}_{\pi}$. Let $\mathbf{S}_{\pi}$ contains $X$'s each child $S$ where arc $X\rightarrow S$ is a segment of a path in $\pi$; let $\bar{\mathbf{S}}_{\pi}$ contains $X$'s each child $S$ where either $S$ is not included in any path from $C$ to $E$, or arc $X\rightarrow S$ is a segment of a path not in $\pi$. Finally, replace values $x_{1}$ with $x_{2}$ for the terms corresponding to nodes in $\mathbf{S}_{\pi}$, and keep values $x_{1}$ unchanged for the terms corresponding to nodes in $\bar{\mathbf{S}}_{\pi}$.

\begin{figure}
	\centering
		\includegraphics[width=1.5in]{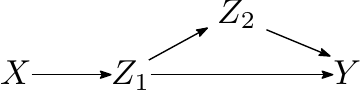}
	\caption{An example with the recanting witness criterion satisfied.}
	\label{fig:rwc}
\end{figure}

Note that the above computation requires $\mathbf{S}_{\pi}\cap \bar{\mathbf{S}}_{\pi} = \emptyset$. Theorem \ref{thm:rwc} is reflected in that: $\mathbf{S}_{\pi}\cap \bar{\mathbf{S}}_{\pi} \neq \emptyset$ if and only if the recanting witness criterion for the $\pi$-specific effect is satisfied. Figure \ref{fig:rwc} shows an example with the recanting witness criterion satisfied, where $\pi = \{X\rightarrow Z_{1}\rightarrow Z_{2}\rightarrow Y\}$. According to the definitions, $Z_{1}$ is contained in both $\mathbf{S}_{\pi}$ and $\bar{\mathbf{S}}_{\pi}$.



\section{Modeling Direct and Indirect Discrimination as Path-Specific Effects}
Consider a historical dataset $\mathcal{D}$ that contains a group of tuples, each of which describes the profile of an individual. Each tuple is specified by a set of attributes $\mathbf{V}$, including the protected attributes, the decision, and the non-protected attributes. Among the non-protected attributes, assume there is a set of attributes that cannot be objectively justified if used in the decision making process, which we refer to as the \emph{redlining attributes} denoted by $\mathbf{R}$. For ease of presentation, we assume that there is only one protected attribute with binary values. We denote the protected attribute by $C$ associated with two domain values $c^{-}$ (e.g., female) and $c^{+}$ (e.g., male); denote the decision by $E$ associated with two domain values $e^{-}$ (i.e., negative decision) and $e^{+}$ (i.e., positive decision). Our approach can extend to handling multiple domain values of $C$ and even multiple $C$s. We assume that a causal graph $\mathcal{G}$ can be built to correctly represent the causal structure of dataset $\mathcal{D}$. In the past decades, many algorithms have been proposed to learn the causal network from data and they are proved to be quite successful \cite{spirtes2000causation,neapolitan2004learning,colombo2014order,kalisch2007estimating}. We also make a reasonable assumption that $C$ has no parent in $\mathcal{G}$, as the protected attribute is always an inherent nature of an individual. 

Discrimination can be captured by the causal effect of $C$ on $E$. In our context, the causal effect includes direct/indirect discrimination and other explainable effects. We model direct discrimination as the causal effect transmitted along the direct path from $C$ to $E$, i.e., $C\rightarrow E$. Define $\pi_{d}$ as the path set that contains only $C\rightarrow E$. The $\pi_{d}$-specific effect of the change of $C$ from $c^{-}$ to $c^{+}$ on $E=e^{+}$ is given by 
\begin{equation*}
SE_{\pi_{d}}(c^{+},c^{-}) = P(e^{+}\mid do(c^{+}|_{\pi_{d}})) - P(e^{+}\mid do(c^{-})). 
\end{equation*}
The physical meaning of $SE_{\pi_{d}}(c^{+},c^{-})$ is the expected change in decisions (in term of the probability of $E=e^{+}$) of individuals from protected group $c^{-}$, if it is told that these individuals were from the other group $c^{+}$ and everything else remains unchanged. When applied to the example in Figure \ref{fig:toy}, it means the expected change in loan approval of applications actually from the disadvantage group (e.g., black), when the bank is instructed to treat the applicants as from the advantage group (e.g., white). Thus, the measurement of the $\pi_{d}$-specific effect exactly follows the definition of direct discrimination and is appropriate for measuring the discriminatory effect. 

Similarly, we model indirect discrimination as the causal effect transmitted along all the indirect paths from $C$ to $E$ that contain the redlining attributes. Given the set of redlining attributes $\mathbf{R}$, define $\pi_{i}$ as the path set that contains all the causal paths from $C$ to $E$ which pass through $\mathbf{R}$, i.e., each of the paths includes at least one node in $\mathbf{R}$. Then, the $\pi_{i}$-specific effect, which is given by 
\begin{equation*}
SE_{\pi_{i}}(c^{+},c^{-}) = P(e^{+}\mid do(c^{+}|_{\pi_{i}})) - P(e^{+}\mid do(c^{-})), 
\end{equation*}
represents the expected change in decisions of individuals from protected group $c^{-}$, if the profiles of these individuals along path $\pi_{i}$ were changed as if they were from the other group $c^{+}$. When applied to the example in Figure \ref{fig:toy}, it means the expected change in loan approval of the disadvantage group if they had the same racial makeups shown in the Zip code as the advantage group. Thus, the $\pi_{i}$-specific effect is appropriate for measuring the discriminatory effect of indirect discrimination. 

Therefore, we propose the criterion for claiming direct and indirect discrimination based on the path-specific effect. We say that direct discrimination against protected group $c^{-}$ is claimed if $SE_{\pi_{d}}(c^{+},c^{-})>\tau$, where $\tau>0$ is a use-defined threshold for discrimination depending on the law. For instance, the 1975 British legislation for sex discrimination sets $\tau=0.05$, namely a 5\% difference. Similarly, given the redlining attributes $\mathbf{R}$, we say that indirect discrimination against protected group $c^{-}$ is claimed if $SE_{\pi_{i}}(c^{+},c^{-})>\tau$. To avoid reverse discrimination, we do not specify which group is the protected group. Therefore, we give the following criterion.
\begin{theorem}\label{thm:criterion}
Given the protected attribute $C$, decision $E$, and redlining attributes $\mathbf{R}$, direct discrimination is claimed if either $SE_{\pi_{d}}(c^{+},c^{-})>\tau$ or $SE_{\pi_{d}}(c^{-},c^{+})>\tau$ holds, and indirect discrimination is claimed if either $SE_{\pi_{i}}(c^{+},c^{-})>\tau$ or $SE_{\pi_{i}}(c^{-},c^{+})>\tau$ holds.
\end{theorem}

It is worth noting that, if a path set $\pi$ contains all causal paths from $C$ to $E$, it can be directly obtained from the definition that the $\pi$-specific effect is equivalent to the total causal effect, i.e., 
\begin{equation*}
SE_{\pi}(c^{+},c^{-}) = TE(c^{+},c^{-}) = P(e^{+}|do(c^{+}))-P(e^{+}|do(c^{-})).
\end{equation*}
It can be proved straightforwardly using Equation \eqref{eq:do} that the above equation equals to $P(e^{+}|c^{+})-P(e^{+}|c^{-})$, which is known as \emph{risk difference} \cite{romei2014multidisciplinary} widely used for discrimination measurement in the anti-discrimination literature. Therefore, the path-specific effect can be considered as a significant extension to risk difference for explicitly distinguishing the discriminatory effects of direct and indirect discrimination from the total causal effect. On the other hand, we do not necessarily have $SE_{\pi_{d}}(c^{+},c^{-})+SE_{\pi_{i}}(c^{+},c^{-})=SE_{\pi_{d}\cup \pi_{i}}(c^{+},c^{-})$. This implies that there might not be a linear connection between direct and indirect discrimination.


According to Definition \ref{def:rwc} and Theorem \ref{thm:rwc}, it is guaranteed that the recanting witness criterion for the $\pi_{d}$-specific effect is not satisfied since there is no intermediate node in the direct path $C\rightarrow E$ and $\mathbf{S}_{\pi_{d}}$ contains $E$ only, and direct discrimination can always be measured from the observational data. Thus, $SE_{\pi_{d}}$ can be computed as follows. 
\begin{equation}\label{eq:sed}
\begin{split}
& SE_{\pi_{d}}(c^{+},c^{-}) = \!\!\! \sum_{\mathbf{V}\backslash\{C,E\}} \bigg(  P(e^{+}|c^{+},Pa(E)\backslash{\{C\}}) \\
& \quad\quad\quad\quad\quad\quad \prod_{V\in \mathbf{V}\backslash\{C,E\}} P(v|Pa(V))\delta_{C=c^{-}} \bigg) - P(e^{+}|c^{-}).
\end{split}
\end{equation}

For indirect discrimination, we divide $C$'s children into $\mathbf{S}_{\pi_{i}}$ and $\bar{\mathbf{S}}_{\pi_{i}}$. Different from above, the recanting witness criterion for the $\pi_{i}$-specific effect might be satisfied or not. When the recanting witness criterion is not satisfied, we have $\mathbf{S}_{\pi_{i}}\cap \bar{\mathbf{S}}_{\pi_{i}} = \emptyset$. Then, $SE_{\pi_{i}}(c^{+},c^{-})$ can be computed as follows.
\begin{equation}\label{eq:sei}
\begin{split}
& SE_{\pi_{i}}(c^{+},c^{-}) = \sum_{\mathbf{V}\backslash\{C\}} \bigg(   \prod_{G\in \mathbf{S}_{\pi_{i}}}P(g|c^{+},Pa(G)\backslash \{C\}) \\
&  \prod_{H\in \bar{\mathbf{S}}_{\pi_{i}}}P(h|c^{-},Pa(H)\backslash \{C\}) \!\!\!\!\!\! \prod_{O\in \mathbf{V}\backslash (\{C\}\cup Ch(C))} \!\!\!\!\!\! P(o|Pa(O))\delta_{C=c^{-}} \bigg) \\
& - P(e^{+}|c^{-}).
\end{split}
\end{equation}

How to deal with the opposite situation will be discussed later in the next section.

\section{Discrimination Discovery and Removal}

\subsection{Discrimination Discovery}
We propose a Path-Specific based Discrimination Discovery (\emph{PSE-DD}) algorithm based on Theorem \ref{thm:criterion}. It first builds the causal network from the historical dataset, and then computes $SE_{\pi_{d}}$ and $SE_{\pi_{i}}$ according to Equations \eqref{eq:sed} and \eqref{eq:sei}. The procedure of the algorithm is shown in Algorithm \ref{alg:ddd}. 

\begin{algorithm}
\SetAlgoVlined
\small
    \SetKwInOut{Input}{Input}
    \SetKwInOut{Output}{Output}
		
		\Input{Historical dataset $\mathcal{D}$, protected attribute $C$, decision attribute $E$, user-defined parameter $\tau$.}
    \Output{Judgment of direct and indirect discrimination $judge_d$, $judge_i$.}
		$\mathcal{G}=buildCausalNetwork(\mathcal{D})$\;
		$judge_d = judge_i = false$\;
		Compute $SE_{\pi_{d}}(\cdot)$ according to Equation \eqref{eq:sed}\;
		\If{$SE_{\pi_{d}}(c^{+},c^{-})> \tau$ $\|$ $SE_{\pi_{d}}(c^{-},c^{+})> \tau$} {
			$judge_d = true$\;
		}
		Compute $SE_{\pi_{i}}(\cdot)$ according to Equation \eqref{eq:sei}\;
		\If{$SE_{\pi_{i}}(c^{+},c^{-})> \tau$ $\|$ $SE_{\pi_{i}}(c^{-},c^{+})> \tau$} {
			$judge_i = true$\;
		}
		\Return{$[judge_d,judge_i]$}\;
    \caption{\emph{PSE-DD}}
		\label{alg:ddd}
\end{algorithm}

The computational complexity of \emph{PSE-DD} depends on the complexities of building the causal network and computing the path-specific effect according to Equation \eqref{eq:sed} or \eqref{eq:sei}.
Many researches have been devoted to improving the performance of network construction \cite{kalisch2007estimating,tsamardinos2003scaling,aliferis2010local} and probabilistic inference in causal networks \cite{heckerman1994new,heckerman1996causal}.
These topics are beyond the scope of this paper.

For indirect discrimination (line 7), the complexity further depends on how to identify $\mathbf{S}_{\pi_{i}}$ and $\bar{\mathbf{S}}_{\pi_{i}}$. 
A straightforward method of finding all paths in $\pi_{i}$ may have an exponential complexity.
On the other hand, it can be easily observed that, a node $S$ belongs to $\mathbf{S}_{\pi_{i}}$ if and only if there exists a path from $S$ to $E$ passing through $\mathbf{R}$ (a path from $S$ to $E$ passing through $\mathbf{R}$ also includes the path where $S$ itself belongs to $\mathbf{R}$). Similarly, $S$ belongs to $\bar{\mathbf{S}}_{\pi_{i}}$ if and only if there does not exist a path from $S$ to $E$ passing through $\mathbf{R}$. It is relatively easy to check the existence of a path between two nodes. In our algorithm, we examine the existence of a path from $S$ to $E$ passing through $\mathbf{R}$ by checking whether there exists a node $R\in \mathbf{R}$ so that $R$ is $S$'s decedent and $E$ is $R$'s decedent. The subroutine of finding $\mathbf{S}_{\pi_{i}}$ and $\bar{\mathbf{S}}_{\pi_{i}}$ is presented in the pseudo-code below, where $De(\cdot)$ denotes the decedents of a node. Since the decedents of all the nodes involved in the algorithm can be obtained by traversing the network starting from $C$ within the time of $O(|\mathbf{A}|)$, the computational complexity of this procedure is given by $O(|\mathbf{V}|^{2}+|\mathbf{A}|)$.

{
\setlength{\interspacetitleruled}{-.4pt}%
\begin{algorithm}
\SetAlgoVlined
\small
		$\mathbf{S}_{\pi_{i}} = \emptyset$, $\bar{\mathbf{S}}_{\pi_{i}} = \emptyset$\;
		\ForEach{$S\in Ch(C)\backslash \{E\}$} {
			\ForEach{$R\in \mathbf{R}$} {
				\eIf{$R\in De(S)\cup \{S\}$ $\&\&$ $E\in De(R)$} {
					$\mathbf{S}_{\pi_{i}} = \mathbf{S}_{\pi_{i}}\cup \{S\}$\;
				} {
					$\bar{\mathbf{S}}_{\pi_{i}} = \bar{\mathbf{S}}_{\pi_{i}}\cup \{S\}$\;
				}
			}
		}		
\end{algorithm}
}

\vspace{-0.4cm}

\subsection{Discrimination Removal}
When direct or indirect discrimination is claimed for a dataset, the discriminatory effects need to be removed before the dataset is released for predictive analysis (e.g., building a classifier). A naive approach would be simply not using the protected attribute when building the predictive model, which often incur significant utility loss. In addition, this approach can eliminate direct discrimination, but indirect discrimination still presents.

We propose a Path-Specific Effect based Discrimination Removal (\emph{PSE-DR}) algorithm to remove both direct and indirect discrimination. The general idea is to modify the causal network and then use it to generate a new dataset. Specifically, we modify the CPT of $E$, i.e., $P(e|Pa(E))$, to obtain a new CPT $P'(e|Pa(E))$, so that the direct and indirect discriminatory effects are below the threshold $\tau$. To maximize the utility of the modified dataset, we minimize the Euclidean distance between the joint distribution of the original causal network (denoted by $P(\mathbf{v})$) and the joint distribution of the modified causal network (denoted by $P'(\mathbf{v})$). As a result, we obtain the following quadratic programming problem.
\begin{equation*}
\begin{split}
\textrm{minimize} & \qquad \sum_{\mathbf{V}}\Big(  P'(\mathbf{v})-P(\mathbf{v}) \Big)^{2} \\
\textrm{subject to} & \qquad SE_{\pi_{d}}(c^{+},c^{-})\leq \tau, \quad SE_{\pi_{d}}(c^{-},c^{+})\leq \tau, \\
& \qquad SE_{\pi_{i}}(c^{+},c^{-})\leq \tau, \quad SE_{\pi_{i}}(c^{-},c^{+})\leq \tau, \\
& \qquad \forall Pa(E),\quad P'(e^{-}|Pa(E))+P'(e^{+}|Pa(E)) = 1, \\
& \qquad \forall Pa(E),e,\quad Pr'(e|Pa(E)) \geq 0,
\end{split}
\end{equation*}
where $P'(\mathbf{v})$ and $P(\mathbf{v})$ are computed according to Equation \eqref{eq:bn} using $P'(e|Pa(E))$ and $P(e|Pa(E))$ respectively, and $SE_{\pi_{d}}(\cdot)$ and $SE_{\pi_{i}}(\cdot)$ are computed according to Equations \eqref{eq:sed} and \eqref{eq:sei} respectively using $P'(e|Pa(E))$. The optimal solution is obtained by solving the quadratic programming problem. After that, the joint distribution of the modified causal network is computed using Equation \eqref{eq:bn}, and the new dataset is generated based on the joint distribution. The procedure of \emph{PSE-DR} is shown in Algorithm \ref{alg:dr}. 

\begin{algorithm}
\SetAlgoVlined
\small
    \SetKwInOut{Input}{Input}
    \SetKwInOut{Output}{Output}
		
		\Input{Historical dataset $\mathcal{D}$, protected attribute $C$, decision attribute $E$, use-defined redlining attributes $\mathbf{R}$, user-defined parameter $\tau$.}
    \Output{Modified dataset $\mathcal{D}^{*}$.}
		Obtain the modified CPT of $E$ by solving the quadratic programming problem\;
		Calculate $P^{*}(\mathbf{v})$ according to Equation (\ref{eq:bn}) using the modified CPTs\;
		Generate $\mathcal{D}^{*}$ based on $P^{*}(\mathbf{v})$\;		
		\Return{$\mathcal{D}^{*}$}\;
		\caption{\emph{PSE-DR}}
		\label{alg:dr}
\end{algorithm}

For discrimination removal, it is crucial to ensure that not only the modified data does not contain discrimination, but also the predictive models built on it will not incur biased decision. The goal of a predictive model is to learn from data the computational relationship between $E$ and all the other attributes, which is captured by the CPT of $E$ in the causal network. In our approach, we modify only the CPT of $E$ to remove discrimination. Therefore, we can ensure that the predictive models can learn these modifications and will not incur discrimination in decision making. We will evaluate this result in the experiments.


The computational complexity of \emph{PSE-DR} depends on the complexity of solving the quadratic programming problem. It can be easily shown that, the coefficients of the quadratic terms in the objective function form a positive definite matrix. According to \cite{kozlov1980polynomial}, the quadratic programming can be solved in polynomial time. Finally, it is also worth noting that our approach can be easily extended to handle the situation where either direct or indirect discrimination needs to be removed.

\subsection{Dealing with Unidentifiable Situation}\label{sec:uni}
As stated in Theorem \ref{thm:rwc}, when the recanting witness criterion is satisfied, the $\pi_{i}$-specific effect cannot be estimated from the observational data. However, the structure of the recanting witness criterion implies indirect discrimination as there exist causal paths from $C$ to $E$ passing through the redlining attributes. From the data owners' perspective, they may want to ensure non-discrimination even though the discriminatory effect cannot be accurately measured. In this case, we remove discrimination by adapting Algorithm \ref{alg:dr} as follows. Recall that $\mathbf{S}_{\pi_{i}}\cap \bar{\mathbf{S}}_{\pi_{i}} \neq \emptyset$ if and only if the recanting witness criterion is satisfied. For each node $S\in \mathbf{S}_{\pi_{i}}\cap \bar{\mathbf{S}}_{\pi_{i}}$, we cut off all the causal paths from $S$ to $E$ that pass through $\mathbf{R}$, so that $S$ would not belong to $\mathbf{S}_{\pi_{i}}$ any more. Then, we must have $\mathbf{S}_{\pi_{i}}\cap \bar{\mathbf{S}}_{\pi_{i}} = \emptyset$ after the modification. 
To cut off the paths, we focus on the arc from $E$'s each parent $Q$, i.e., $Q\rightarrow E$. If these exists a path from $S$ to $Q$ passing through $\mathbf{R}$, then arc $Q\rightarrow E$ is removed from the network. 
The pseudo-code of this procedure is shown below, which can be added before line 1 in Algorithm \ref{alg:dr} to deal with this situation.

{
\setlength{\interspacetitleruled}{-.4pt}%
\begin{algorithm}
\SetAlgoVlined
\small
		\If{$\mathbf{S}_{\pi_{i}}\cap \bar{\mathbf{S}}_{\pi_{i}} \neq \emptyset$} {
			\ForEach{$S\in \mathbf{S}_{\pi_{i}}\cap \bar{\mathbf{S}}_{\pi_{i}}$} {
				\ForEach{$Q\in Pa(E)$} {
					\ForEach{$R\in \mathbf{R}$}{
						\If{$R\in De(S)$ $\&\&$ $Q\in De(R)$} {
							Remove arc $Q\rightarrow E$ from $\mathcal{G}$\;
							Break\;
						}
					}
				}
			}
		}
\end{algorithm}
}

\vspace{-0.4cm}

\section{Experiments}
In this section, we conduct experiments using two real datasets: the Adult dataset \cite{adultdataset} and the Dutch Census of 2001 \cite{dutchdataset}. We compare our algorithms with the local massaging (\emph{LMSG}) and local preferential sampling (\emph{LPS}) algorithms proposed in \cite{zliobaite2011handling} and disparate impact removal algorithm (\emph{DI}) proposed in \cite{feldman2015certifying,Adler2016}.
The causal networks are constructed and presented by utilizing an open-source software TETRAD \cite{tetrad}. We employ the original PC algorithm \cite{spirtes2000causation} and set the significance threshold $0.01$ for conditional independence testing in causal network construction. The quadratic programming is solved using CVXOPT \cite{dahl2006cvxopt}.

\subsection{Discrimination Discovery}

\begin{figure*}[ht]
	\centering
		\includegraphics[width=5.7in]{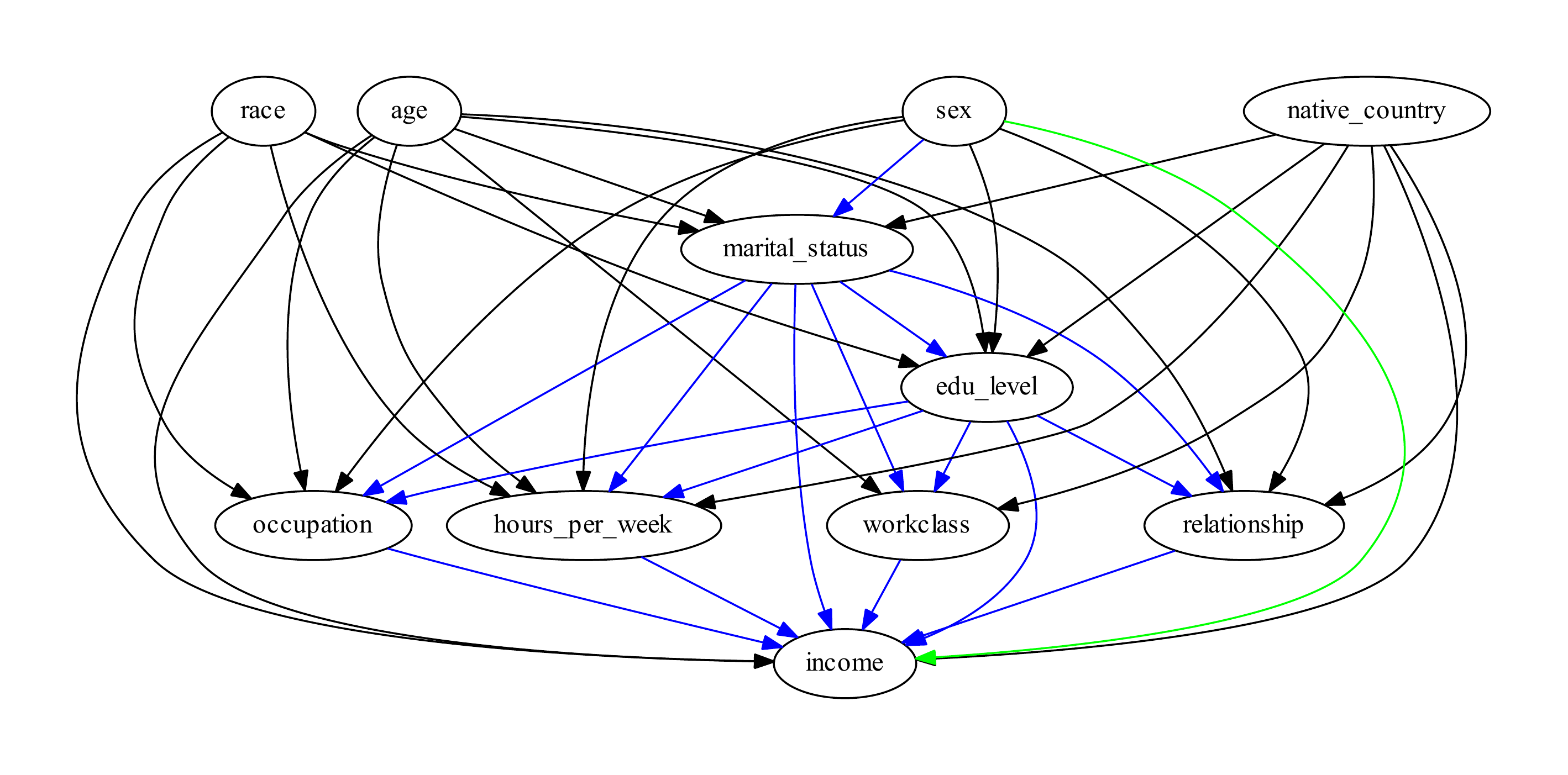}
	\caption{Causal network for Adult dataset: the green path represents the direct path, and the blue paths represent the indirect paths passing through \texttt{marital\_status}.}
	\label{fig:adult2}
\end{figure*}

\begin{figure*}[ht]
	\centering
		\includegraphics[width=6in]{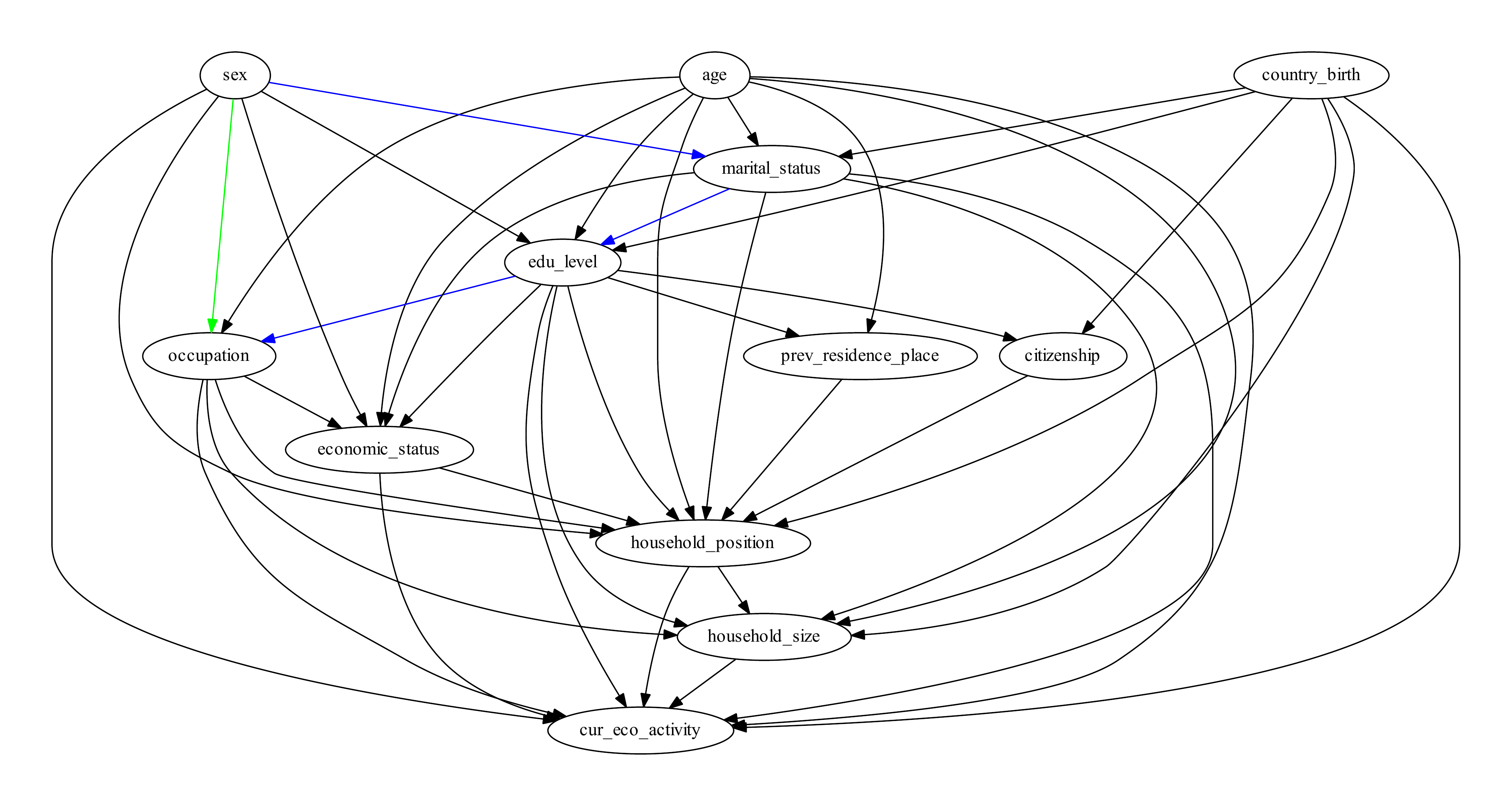}
	\caption{Causal network for Dutch dataset: the green path represents the direct path, and the blue paths represent the indirect paths passing through \texttt{marital\_status}.}
	\label{fig:dutch2}
\end{figure*}

The Adult dataset consists of 65123 tuples with 11 attributes such as \texttt{age}, \texttt{education}, \texttt{sex}, \texttt{occupation}, \texttt{income}, \texttt{marital\_status} etc. Since the computational complexity of the PC algorithm is an exponential function of the number of attributes and their domain sizes, for computational feasibility we binarize each attribute's domain values into two classes to reduce the domain sizes. We use three tiers in the partial order for temporal priority: \texttt{sex}, \texttt{age}, \texttt{native\_country}, \texttt{race} are defined in the first tier, \texttt{edu\_level} and \texttt{marital\_status} are defined in the second tier, and all other attributes are defined in the third tier. The causal network is shown in Figure \ref{fig:adult2}. We treat \texttt{sex} as the protected attribute, \texttt{income} as the decision, and \texttt{marital\_status} as the redlining attribute. The green path represents the direct path from \texttt{sex} to \texttt{income}, and the blue paths represent the indirect paths passing through \texttt{marital\_status}. 
We set the discrimination threshold $\tau$ as 0.05. By computing the path-specific effects, we obtain that $SE_{\pi_{d}}(c^{+},c^{-}) = 0.025$ and $SE_{\pi_{i}}(c^{+},c^{-}) = 0.175$, which indicate no direct discrimination but significant indirect discrimination against females according to our criterion.

The Dutch dataset consists of 60421 tuples with 12 attributes. Similarly, we binarize the domain values of attribute age due to its large domain size. Three tiers are used in the partial order for temporal priority: \texttt{sex}, \texttt{age}, \texttt{country\_birth} are defined in the first tire, \texttt{edu\_level} and \texttt{marital\_status} are defined in the second tire, and all other attributes are defined in the third tire. The causal graph is shown in Figure \ref{fig:dutch2}. Similarly we treat \texttt{sex} as the protected attribute, \texttt{occupation} as the decision, and \texttt{marital\_status} as the redlining attribute. For this dataset, $SE_{\pi_{d}}(c^{+},c^{-}) = 0.220$ and $SE_{\pi_{i}}(c^{+},c^{-}) = 0.001$, indicating significant direct discrimination but no indirect discrimination against females.

\subsection{Discrimination Removal}
We run the removal algorithm \emph{PSE-DR} to remove discrimination from the Adult and Dutch datasets. Then, we run the discovery algorithm \emph{PSE-DD} to further examine whether discrimination is truly removed in the modified datasets. For the modified Adult dataset we have $SE_{\pi_{d}}(c^{+},c^{-}) = 0.013$ and $SE_{\pi_{i}}(c^{+},c^{-}) = 0.049$, and for the modified Dutch dataset we have $SE_{\pi_{d}}(c^{+},c^{-}) = 0.050$ and $SE_{\pi_{i}}(c^{+},c^{-}) = 0.001$. The results show that the modified datasets contain no direct and indirect discrimination.


\textbf{Discrimination in predictive models.} We aim to examine whether the predictive models built from the modified dataset incur discrimination in decision making. We use the Adult dataset where indirect discrimination is detected, and divide the original dataset into the training and testing datasets. First, we remove discrimination from the training dataset to obtain the modified training dataset. Then, we build the predictive models from the modified training dataset, and use them to make predictive decisions over the testing data. Two classifiers, SVM and Decision Tree, are used for prediction with five-fold cross-validation. Finally, we run \emph{PSE-DD} to examine whether the predictions for the testing data contain discrimination. We also examine the data utility ($\chi^{2}$) and the prediction accuracy.

The results are shown in Table \ref{tab:rd}. As shown in the column ``\emph{PSE-DD}'', both the modified training data and the predictions for the testing data contain no direct and indirect discrimination. In addition, \emph{PSE-DD} produces relatively small data utility loss in term of $\chi^2$ and good prediction accuracy. For comparison, we include algorithms from previous works: \emph{LMSG}, \emph{LPS} and \emph{DI}. For \emph{LMSG} and \emph{LPS}, discrimination is not removed even from the training data, and hence also exists in the predictions. The \emph{DI} algorithm provides a parameter $\lambda$ to indicate the amount of discrimination to be removed, where $\lambda=0$ represents no modification and $\lambda=1$ represents full discrimination removal. However, $\lambda$ has no direct connection with the threshold $\tau$. In our experiments, we execute $DI$ multiple times with different $\lambda$s and report the one that is closest to achieve $\tau=0.05$. As shown in the column ``\emph{DI}'', it indeed removes direct and indirect discrimination from the training data. However, as indicated by the bold values 0.167/0.168, significant amount of indirect discrimination exists in the predictions of both classifiers. In addition, its data utility is far more worse than \emph{PSE-DR}, implying that it removes many information unrelated to discrimination.

\setlength{\tabcolsep}{1.8pt}
\begin{table}[tbp]
	\centering
	\caption{Direct/indirect discriminatory effects in the modified training data and predictions for the testing data. Values violating the discrimination criterion are marked in bold. }
	\label{tab:rd}
	\begin{tabular}{|c|c|c|c|c|c|c|}
		\Xhline{0.75pt}
		\multicolumn{3}{|c|}{}                                                       & \textit{PSE-DD}  &  \textit{DI}   &  \textit{LMSG}  &  \textit{LPS}   \\ \Xhline{0.75pt}
		\multicolumn{2}{|c|}{\multirow{3}{*}{Train}}      &        Direct         & 0.013 & 0.001 & -0.142 & -0.142 \\ \cline{3-7}
		\multicolumn{2}{|c|}{}                               &       Indirect        & 0.049 & 0.050 & \textbf{0.288}  & \textbf{0.174}  \\ \cline{3-7}
		\multicolumn{2}{|c|}{}                               & $\chi^2(\times 10^4)$ & 1.620 & 7.031 & 1.924  & 1.292  \\ \Xhline{0.75pt}
		\multirow{6}{*}{Predict} &  \multirow{3}{*}{SVM}  &        Direct         & 0.023 & 0.005 & -0.124 & -0.051 \\ \cline{3-7}
		&                        &       Indirect        & 0.041 & \textbf{0.167} & \textbf{0.271}  & \textbf{0.192}  \\ \cline{3-7}
		&                        &       Accu.(\%)        & 80.54 & 81.47 & 76.81  & 76.63  \\ \cline{2-7}
		& \multirow{3}{*}{DTree} &        Direct         & 0.023 & 0.004 & -0.124 & -0.051 \\ \cline{3-7}
		&                        &       Indirect        & 0.042 & \textbf{0.168} & \textbf{0.271}  & \textbf{0.192}  \\ \cline{3-7}
		&                        &       Accu.(\%)        & 80.55 & 81.38 & 76.81  & 76.64  \\ \Xhline{0.75pt}
	\end{tabular}
\end{table}

\section{Related Work}
A number of techniques have been proposed to discover discrimination in the literature. 
Classification rule-based methods such as \emph{elift} \cite{pedreshi2008discrimination} and \emph{belift} \cite{mancuhan2014combating} were proposed to represent certain discrimination patterns. \cite{luong2011k,zhang2016situation} dealt with the individual discrimination by finding a group of similar individuals. \cite{zliobaite2011handling} proposed conditional discrimination which considers some part of the discrimination may be explainable by certain attributes. None of these work explicitly identifies direct discrimination, indirect discrimination, and explainable effects. In \cite{DBLP:journals/corr/BonchiHMR15}, the authors proposed a framework based on the Suppes-Bayes causal network and developed several random-walk-based methods to detect different types of discrimination. However, the construction of the Suppes-Bayes causal network is impractical with the large number of attribute-value pairs. In addition, it is unclear how the number of random walks is related to practical discrimination metrics, e.g., the difference in positive decision rates.


Proposed methods for discrimination removal are either based on data preprocessing \cite{kamiran2012data,zliobaite2011handling} or algorithm tweaking \cite{kamiran2010discrimination,calders2010three,kamishima2011fairness}. 
In a recent work \cite{feldman2015certifying}, the authors first ensure no direct discrimination by completely removing the protected attribute $C$ from data, and then modify all the non-protected attributes to ensure that $C$ cannot be predicted from the non-protected attributes. As a result, indirect discrimination is removed since the decision $E$ has no connection with $C$ via the non-protected attributes. However, as shown in our experiment results, this approach cannot ensure that predictions made by the classifier built on the modified data do not contain discrimination. In addition, it suffers significant utility loss as it removes all the connections between $C$ and $E$.

\section{Conclusions}
In this paper, we studied the problem of discovering both direct and indirect discrimination from the historical data, and removing the discriminatory effects before performing predictive analysis. We made use of the causal network to capture the causal structure of the data, and modeled direct and indirect discrimination as different path-specific effects. Based on that, we proposed the discovery algorithm \emph{PSE-DD} to discover both direct and indirect discrimination, and the removal algorithm \emph{PSE-DR} to remove them. The experiments using real datasets show that, only our approach can ensure that the predictive models built from the modified data are not subject to any type of discrimination.


\bibliographystyle{aaai}
\bibliography{paper}

\end{document}